\documentclass{article}

\usepackage{arxiv}

\usepackage{amsmath}
\usepackage{multirow}
\usepackage[utf8]{inputenc} 
\usepackage[T1]{fontenc}    
\usepackage{hyperref}       
\usepackage{url}            
\usepackage{booktabs}       
\usepackage{amsfonts}       
\usepackage{nicefrac}       
\usepackage{microtype}      
\usepackage{lipsum}
\usepackage{graphicx}
\graphicspath{ {./images/} }

\title{ADEPT: Adaptive Dynamic Early-Exit Process for Transformers}

\author{
 Sangmin Yoo\thanks{Internship at Samsung Semiconductor, San Jose, CA, USA} \\
  University of Michigan\\
  Ann Arbor, MI, USA\\
  \texttt{ysangmin@umich.edu} \\
  \And
 Srikanth Malla \\
  Samsung Semiconductor\\
  San Jose, CA, USA\\
  \texttt{srikanth.m@samsung.com} \\
  \And
 Chiho Choi \\
  Samsung Semiconductor\\
  San Jose, CA, USA\\
  \texttt{chiho1.choi@samsung.com} \\
  \And
 Wei D. Lu \\
  University of Michigan\\
  Ann Arbor, MI, USA\\
  \texttt{wluee@umich.edu} \\
  \And
 Joon Hee Choi\thanks{Corresponding author} \\
  Samsung Semiconductor\\
  San Jose, CA, USA\\
  \texttt{jh4.choi@samsung.com} \\
}

\begin{document}

\footnotetext{These authors contributed equally: Sangmin Yoo, Srikanth Malla.}

\maketitle
\begin{abstract}
The inference of large language models imposes significant computational workloads, often requiring the processing of billions of parameters. Although early-exit strategies have proven effective in reducing computational demands by halting inference earlier, they apply either to only the first token in the generation phase or at the prompt level in the prefill phase. Thus, the Key-Value (KV) cache for skipped layers remains a bottleneck for subsequent token generation, limiting the benefits of early exit. We introduce ADEPT (\textbf{A}daptive \textbf{D}ynamic \textbf{E}arly-exit \textbf{P}rocess for \textbf{T}ransformers), a novel approach designed to overcome this issue and enable dynamic early exit in both the prefill and generation phases. The proposed adaptive token-level early-exit mechanism adjusts computation dynamically based on token complexity, optimizing efficiency without compromising performance. ADEPT further enhances KV generation procedure by decoupling sequential dependencies in skipped layers, making token-level early exit more practical. Experimental results demonstrate that ADEPT improves efficiency by up to 25\% in language generation tasks and achieves a $4\times$ speed-up in downstream classification tasks, with up to a 45\% improvement in performance.
\end{abstract}


\section{Introduction}
Transformer~\cite{attention_is_all_you_need} has attracted tremendous attention as a foundational building block for large language models (LLMs). By leveraging its scalability and contextual understanding, multi-functional models like GPT3~\cite{gpt3-original}, LLAMA~\cite{llama-original}, PALM~\cite{PALM} have been developed as general-purpose pre-trained LLMs. Adapting these models through finetuning has become a standard practice across various natural language processing tasks, from language translation~\cite{fine-tune-language-translation1,fine-tune-language-translation2} and sentiment analysis~\cite{fine-tune-sent-anal1,fine-tune-sent-anal2} to question-answering~\cite{fine-tune-QnA1,fine-tune-QnA2} and text generation~\cite{fine-tune-text-gen1, fine-tune-text-gen2}. Despite notable progress, LLMs's complex architectures and an enormous number of trainable parameters make the finetuning and inference of the models unavailable in limited computational environments. Even in computationally rich settings, these heavy-weight LLMs often encounter issues called \textit{analysis paralysis} (or \textit{overthinking}) during inference~\cite{shallow-deep} where early layers in the models can capture sufficient information for proper predictions. In such cases, the quality of information can be marginally improved throughout the following layers, resulting in poor cost-effectiveness.

For cost-effectiveness, many efforts have been made in the area of early-exit~\cite{shallow-deep, pabee, fpabee, deebert, distilbert, branchynet, berxit, CALM, FREE, mue}. They enable LLMs to generate outputs in the middle of the recurrent layers while keeping the comparable performance to models that complete the whole layers. However, they commonly leave room for further improvement based on the reasons: (i) current methods are solely dependent on prompt's last token during the prefill phase, thus leaving room for the further performance-efficiency optimization of the rest tokens; (ii) they train additional modules, each dedicated to a layer, or even finetune with entire LLMs, causing an excessive increase in training cost.

We propose ADEPT (\textbf{A}daptive \textbf{D}ynamic \textbf{E}arly-exit \textbf{P}rocess for \textbf{T}ransformers) to address the limitations of existing early-exit strategies, particularly the bottleneck of generating Key-Value (KV) caches for skipped layers, which restricts efficient token-level early exits. ADEPT introduces a decoupled, non-sequential mechanism for processing skipped layers, allowing computations in these layers to be parallelized. 
Furthermore, it eliminates the need to  compute queries, attention, and feedforward blocks for tokens that have exited early. This significantly reduces computational overhead while preserving the dependencies required for accurate processing. Unlike previous methods that rely on costly operations outside the standard model or require resource-intensive fine tuning, ADEPT evaluates each token’s readiness to exit using a lightweight similarity-based criterion. The exit decisions are refined through a shared add-on module, and KV representations are dynamically generated to ensure compatibility with downstream layers. By extending early-exit strategies across both the prefill and generation phases, ADEPT achieves efficient, token-specific computation while maintaining high performance, setting a new standard for scalability and optimization in Transformer models.

By achieving an order-of-magnitude in power efficiency, ADEPT outperforms existing techniques. Notably in a language modeling task, ADEPT achieves the same perplexity score with 20\% computation reduction as the pre-trained base LLM without fine-tuned, while other methods experience performance degradation by 225\% and 216\% in perplexity on the same computational reduction.

Our contributions are threefold:
\begin{itemize}  
   \item We identify limitations in existing early-exit methods, particularly their inability to perform token-level early exit effectively due to sequential dependencies in KV caching, and propose a solution that generalizes early-exit to both prefill and generation phases.  
   \item We introduce a novel early-exit framework for Transformers, formulated as a Markov Decision Process with preference-based rewards, balancing computational efficiency and task performance while addressing key-value dependency constraints in skipped layers.  
   \item We empirically demonstrate that ADEPT outperforms existing methods in both efficiency and performance, achieving significant computational savings in Transformer-based language generation and downstream classification tasks, as well as in multimodal models.  
\end{itemize}

\section{Related Works}
\label{Sec-Related Works}

Recent years have witnessed significant efforts in accelerating LLMs based on Transformer architecture. These efforts have reduced latency (or computational demands) while minimizing performance degradation. DeeBERT~\cite{deebert} inspired by BranchyNet~\cite{branchynet} introduces an early-exit approach based on the entropy of the prediction probability distribution. To get the distribution of the intermediate layers, it incorporates classifier modules, each embedded in an encoder layer in the Transformer model. Addressing intermittent performance drop caused by the prediction probability-based methods, PABEE~\cite{pabee} monitors the number of consecutive identical predictions made by the classifiers. When it reaches the predetermined number, early-exit is triggered. However, PABEE often suffers from its strict cross-layer comparison. It provokes its extension, F-PABEE~\cite{fpabee} based on a similarity between the output distributions of the classifiers.

The models' early-exit operating module in every layer inevitably burdens computational overhead during inference. To mitigate the cost, MuE \cite{mue} proposes an approach relying on a similarity score between hidden states of neighboring layers without the modules. Instead, it employs an additional loss function to finetune the entire base LLM, encouraging the generation of similar hidden states across neighboring layers. While MuE demonstrates out-performance, the finetuning considering all intermediate hidden states incurs substantial training costs. FREE~\cite{FREE} based on a Shallow-Deep module~\cite{shallow-deep} has two exit points and reduces latency by synchronizing decoding of early-exited tokens in parallel when they need to be processed for following token processing. However, it also requires finetuning for the shallower exit with knowledge distillation as an additional loss term, increasing cost as in MuE. Besides, CALM~\cite{CALM} proposes a decaying threshold function of early-exit criteria along layers to enable finer performance-efficiency optimization.

Note that ADEPT determines early-exit at a token level in both prefill and generation phases, while the others do at a prompt level in the prefill phase, or only first generation token because of a fundamental problem of sequential dependency of KV cache in attention computation.

\section{Problem Formulation: Markov Decision Process}

The task of dynamically determining the computation depth for each token in a Transformer model can be framed as a Markov Decision Process (MDP). Each token's trajectory through the model's layers is treated as a sequential decision process, where the goal is to optimize computational efficiency while adhering to task-specific performance constraints.

\paragraph{State Space:}
To formulate our problem, we define each state of the state space $s_{i,j}$ as follows:
\begin{equation}
    s_{i,j} = (s_{i,j}^h, s_{i,j}^{q}, s_{i,j}^{k}, s_{i,j}^{v})
\end{equation}
where \(s_{ij}^h \in \mathbb{R}^{D_h}\), \(s_{i,j}^q, s_{i,j}^k, s_{i,j}^v \in \mathbb{R}^{D_q}\) denote the hidden state, query, key, and value vectors of the $j$-th token at the $i$-th layer, respectively.  \(D_h\) denotes the dimensionality of the hidden state and \(D_q\) represents that of the query, key, and value vector. Also, $i \in [1, N]$ and $j \in [1, M]$ where \(N\) and \(M\) represent the total number of layers and the maximum token sequence length, respectively.

\paragraph{Action Space:}
In the context of early exiting frameworks, at each state \(s_{i,j}\), the action \(a_{i,j} \in \{\text{``Continue"}, \text{``Exit"}\}\) determines whether to continue processing or terminate in early exiting domains.

\paragraph{Preference-Based Reward Formulation:}
Direct Preference Optimization (DPO) naturally extends the framework by aligning the reward function with human or task-specific preferences. In the context of ADEPT, the preference for early exits reflects the goal of minimizing unnecessary computation while maintaining task performance.

The probability of preference \(\mathcal{P}_\pi(a_{i, j} | s_{i, j})\) is modeled on the action \(a_{i, j}\) in the state \(s_{i, j}\), where \(\pi\) represents the policy.
This preference probability is parameterized to balance task-specific performance $\mathcal{T}$ and computational cost $\mathcal{C}$, \(
    \mathcal{P}_\pi(a_{i,j} | s_{i, j}) \propto \exp\left(\lambda \cdot \mathcal{T}_{i, j} - \beta \cdot \mathcal{C}_{i, j}\right), \)
where \(\lambda > 0\) and \(\beta > 0\) are scaling factors controlling the relative importance of performance and cost, respectively. The proof of this formulation is given in Appendix~\ref{supsec:proof_formulation}.

The reward function is defined as the log-likelihood of the preference probability, $
    R(s_{i, j}, a_{i, j}) = \log \mathcal{P}_\pi(a_{i, j} | s_{i, j})$. Then, 
the policy \(\pi^*\) is optimized to maximize the expected cumulative reward \(R(s_{i, j}, a_{i, j})\), integrating over the preference probabilities:
\begin{equation}
\begin{split}
    \pi^* &=\arg\max_\pi \;\; \mathbb{E}_\pi \left[\sum_{i=1}^N \sum_{j=1}^M R(s_{i, j}, a_{i, j})\right]\\
    &= \arg\max_\pi \;\; \mathbb{E}_\pi \left[\sum_{i=1}^N \sum_{j=1}^M \log \mathcal{P}_\pi(a_{i, j} | s_{i, j})\right].    
\end{split}
\end{equation}

\paragraph{Objective:}
The unified objective in Direct Preference Optimization (DPO) incorporates both preference-based rewards and constraints on computational budget and task-specific performance. Therefore, the objective is expressed as:
\begin{equation}
    \mathcal{L}(\pi) = \sum_{i=1}^N \sum_{j=1}^M \log \mathcal{P}_\pi(a_{i,j} | s_{i,j})
\end{equation}
subject to the following constraints:
\begin{equation}
\mathbb{E}_\pi \left[\sum_{i=1}^N \sum_{j=1}^M \mathcal{C}_{i,j}\right] \leq \mathcal{C}_{\text{threshold}},  \mathbb{E}_\pi \left[\sum_{i=1}^N \sum_{j=1}^M \mathcal{T}_{i,j}\right] \geq \mathcal{T}_{\text{min}},
\end{equation}
where \(\mathcal{C}_{\text{threshold}}\) denotes the maximum allowable computational budget, and \(\mathcal{T}_{\text{min}}\) represents the minimum required task-specific performance, providing a principled foundation for the early-exit decisions.

\paragraph{Sequential Dependency in Attention:}
Transitions between hidden states are central to the Transformer layer's operations. These transitions, defined through the layer's operations, determine how information flows through the model. Transitions between hidden states are governed by the Transformer layer's operations.

The sequential dependency in the Transformer model ensures that the transition probability for the hidden state \(s_{i, j}^h\) at layer \(i\) depends not only on the previous hidden state \(s_{i-1,j}^h\) and the layer's parameters \(W_i\), but also on the aggregated key-value representations of all preceding tokens \(s_{i,j}^{kv}\) at the \(i\)-th layer for the \(j\)-th token, where \(s_{i,j}^{kv} \in \mathbb{R}^{D_q \times 2}\) as a short-hand stacked notation of key, and value representations. Formally, given layer's weights \(W_i\), this dependency is expressed as:
\begin{equation}
    P(s_{i, j}^h | s_{i-1,j}^h, \{s_{i,l}^{kv}\}_{l=1}^{j-1}, W_{i}).
\end{equation}
This reflects the fact that the computation of \(s_{i, j}^h\) cannot be decoupled from the representations of earlier tokens in the sequence due to the nature of self-attention. Additionally, given \(W_i^{kv}\) represents the key-value transformation weights, the earlier tokens \(s_{i,j}^{kv}\) depends on a sequential transition \(P(s_{i,j}^{kv} |s^{h}_{i-1,j}, W_i^{kv}) \) which hinders with early-exit benefit. Because of this reason, prior early-exit methods, such as PABEE, F-PABEE, and DeeBERT, could make early-exit decisions to only the first token of the generation phase without any early exit in prefill, and methods like FREE extend to only prompt level early exit in the prefill phase. They halt processing for all tokens simultaneously once a decision is made and cannot perform token level early exit.

\section{Methodology}
\label{Sec-Methodology}
ADEPT enables smooth transitions for early-exited tokens while maintaining attention consistency. By leveraging state convergence, it maps the hidden state at the exit layer to an approximated final state, \(\tilde{s}_{Nj}^h\), ensuring compatibility with later layers. Skipped key-value states are reconstructed as:  
\begin{equation}
\label{eq:decouple_seq_dep}
    P(s_{r,j}^{kv} | \tilde{s}_{N,j}^h, W_r^{kv}) \quad \forall r > i^*,
\end{equation}  
where \(i^*\) is the exited layer. This formulation decouples key-value dependencies \(P(s_{r,j}^{kv} |s^{h}_{r-1,j}, W_r^{kv}) \), eliminating sequential constraints that hinder early exit benefit. As a result, ADEPT enables independent token-level exits in both prefill and generation phases, optimizing efficiency while preserving performance.

\begin{figure}[!t]
\begin{center}
\centerline{\includegraphics[width=0.45\columnwidth]{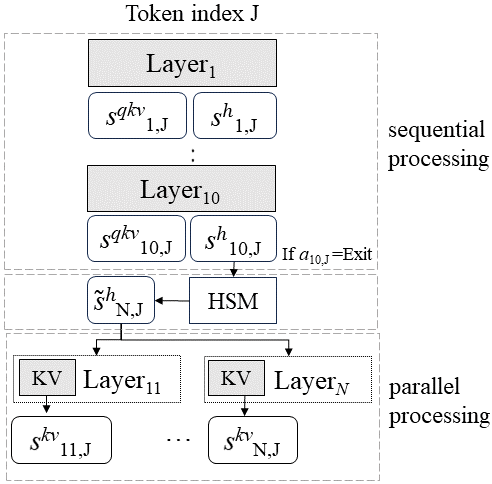}}
\caption{Illustration of early-exit on the token-by-token basis in the prefill. Anticipated last hidden states ($\Tilde{s}_{Nj}^h$) generated by the hidden state mapper (HSM) are propagated to the following layers for the Keys-Values (KVs) generation and the end of the layer chain as the final hidden states. }
\label{fig:token-by-token}
\end{center}
\vspace{-5mm}
\end{figure}

The overview is shown in  Figure~\ref{fig:token-by-token}, which illustrates the workflow of ADEPT during the prefill phase, where tokens dynamically determine their computation depth based on the early-exit criterion. For each token, the decision to exit is governed by the alignment of its current hidden state with accumulated historical representations, enabling token-specific levels of analysis. Upon early exit, the final hidden state, denoted as \(\Tilde{s}_{Nj}^h\), is predicted using a state transformation module (HSM). The skipped layers' key-value vectors are then generated by propagating \(\Tilde{s}_{Nj}^h\) through the respective KV weight matrices, ensuring consistent processing across all tokens, without any sequential unfolding across skipped layers.

\subsection{Token-level Early-Exit Policy} \label{sec:token-level early-exit}
In this work, we address the prior mentioned challenge by token level exit policy motivated by state convergence via clustering behavior in transformers~\cite{geshkovski2024emergence, phang2021fine}.

\textbf{State Convergence Metric:} As the states evolve, they accumulate inertia, reflecting their stabilization over successive layers. To model this, we define the cumulative state \(s_{\text{intertia}, j}^h \in \mathbb{R}^{D_h}\), which is computed as the sum of the normalized hidden states from all preceding layers:
\begin{equation}
    s_{\text{intertia}, j}^h = \sum_{d=1}^{i-1} \hat{s}_{d,j}^h,
\end{equation}
where \(\hat{s}_{d,j}^h = \frac{s_{d,j}^h}{\|s_{d,j}^h\|}\) represents the normalized form of \(s_{d,j}^h\). 
To quantify the alignment between \(s_{i,j}^h\) and \(s_{\text{intertia}, j}^h\), we define the readiness of early exit of tokens with the similarity computed using inner product. 
As a result, a token $j$ is considered ready to exit early at layer \(i\) when:
\begin{equation}
    \pi(a_{i,j}=\text{exit} | s_{ij}^h) = \langle \hat{s}_{i,j}^h, \hat{s}_{\text{intertia}, j}^h \rangle \geq \delta,
\end{equation}
The threshold \(\delta \in (0,1)\) ensures alignment between a token’s current state and its cumulative history, balancing efficiency and model quality. Higher \(\delta\) enforces stricter exits by requiring stronger convergence. Calibrated to match the computational budget \(\mathcal{C}_{\text{threshold}}\), this policy enables ADEPT to dynamically adjust computation depth per token complexity.

\subsection{Hidden State Mapper Architecture}
In early-exit strategies, tokens must maintain a task-specific quality \(\mathcal{T}_{\text{min}}\) to preserve downstream performance. Without proper mapping, their hidden states may result in task-specific quality that falls below this threshold. The Hidden State Mapper (HSM) transforms early-exited token states to approximate final hidden states, ensuring task objectives are met while maintaining efficiency.

Let the current hidden state at layer \(i\) be \(s_{i,j}^h \in \mathbb{R}^{D_h}\), the previous layer's hidden state \(s_{i-1,j}^h \in \mathbb{R}^{D_h}\), and the cumulative inertia state \(s_{\text{inertia}, j}^h \in \mathbb{R}^{D_h}\). In the following, we describe how HSM generates the predicted final hidden state \(\Tilde{s}_{ij}^h \in \mathbb{R}^{D_h}\).

\textbf{State Transformation Pipeline:} The HSM refines input states by first concatenating the current hidden state \(s_{i,j}^h\), the previous hidden state \(s_{i-1,j}^h\), and the inertia state \(s_{\text{inertia},j}^h\), followed by RMS normalization for consistent scaling. The normalized input undergoes a 3x3 convolution to capture local interactions, enhancing feature expressiveness. An MLP with skip connections further processes the convolution output, preserving critical information from the original input. The final predicted hidden state, \(\Tilde{s}_{Nj}^h \in \mathbb{R}^{D_h}\), aligns seamlessly with the Transformer architecture. Detailed architecture is shown in Appendix Section~\ref{supsec:HSM details}.

\subsection{Key and Value Generation} \label{sec:KV GEN}

The HSM is a key component of ADEPT, designed to enable efficient token-level early exits while maintaining compatibility with the Transformer's computations in skipped layers. For tokens that exit early, the HSM generates a final mapped state \(\tilde{s}_{N,j}^h \in \mathbb{R}^{D_h}\), which approximates the hidden states at the final layer \(N\), ensuring consistency and expressiveness. This mapped state \(\tilde{s}_{N,j}^h\) is utilized to compute the key-value representations (\(s_{r,j}^{kv}\)) required for the skipped layers (\(r > i^*\)), , where \(i^*\) denotes the index of the layer where the token exited.

For each skipped layer \(r > i^*\), the propagated state is defined as \(\tilde{s}_{N,j}^h\). The key-value representation for layer \(r\), denoted by \(s_{r,j}^{kv} \in \mathbb{R}^{D_q \times 2}\), is computed using a shared weight matrix \(W_r^{kv} \in \mathbb{R}^{D_h \times (2D_q)}\):
\begin{equation}
s_{r,j}^{kv} = \tilde{s}_{N,j}^h W_r^{kv}.
\end{equation}
This design aligns with the sequential dependency in Transformers, where \(s_{i,j}^h\) computation depends on both the current state \(s_{i-1,j}^h\) and aggregated key-value representations of earlier tokens. By utilizing \(\tilde{s}_{N,j}^h\) for KV computations across skipped layers, ADEPT preserves these dependencies without redundant attention or feedforward computations. Additionally, with decoupled or non-sequential behavior, we can parallelize \(s^{kv}_{r,j}\) as shown in Equation~\ref{eq:decouple_seq_dep}.

In contrast to prior methods like PABEE, F-PABEE, DeeBERT, or FREE—which restrict early-exit decisions to only first-token or prompt-level scenarios—ADEPT dynamically adapts the computation depth for each token. This token-level flexibility, combined with the parallelized KV computations across skipped layers, provides significant improvements in both efficiency and scalability while maintaining high representational fidelity.

\textbf{Task-Specific KV Generation:} ADEPT tailors its early-exit strategy to the specific needs of different tasks, optimizing efficiency further while maintaining performance.

For \textbf{autoregressive tasks} like language modeling and text generation, ADEPT generates key-value representations (\(s_{r,j}^{kv}\)) for skipped layers (\(r > i^*\)) using the mapped state \(\tilde{s}_{N,j}^h\). This ensures that dependencies required for sequential decoding are preserved, while redundant computations, such as attention and feedforward operations, are avoided.

For \textbf{downstream tasks} such as text classification, computations stop entirely beyond the exited layer \(i^*\). The mapped state \(\tilde{s}_{N,j}^h\) is directly used as input for the classifier, eliminating unnecessary processing while preserving task-relevant features.

The task-specific optimization strategies streamline computation by prioritizing essential operations and ensuring resources are utilized solely for contributions that directly impact the model's output.
\begin{figure}[!t]
\begin{center}
\centerline{\includegraphics[width=0.5\columnwidth]{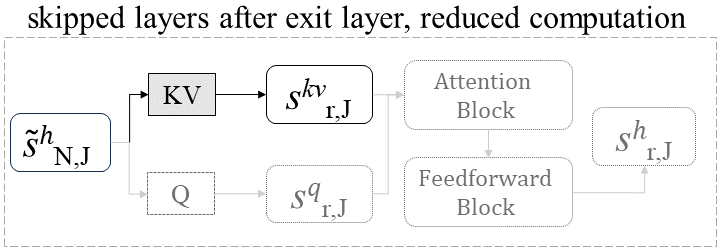}}
\caption{reduced computation by avoid query, attention and feed forward block in the skipped layers (index r \textgreater \space $i^*$), where the exit layer index $i^*$}
\label{fig:reduced_computation}
\end{center}
\vspace{-5mm}
\end{figure}

\section{Experiments} \label{sec:Experiments}

\textbf{Setup:} Experiments were conducted using pre-trained models from Huggingface Transformers~\cite{huggingface}, implemented with PyTorch~\cite{pytorch} and Huggingface Accelerate~\cite{accelerate} on NVIDIA HGX H100 and DGX A100 GPU servers. The pre-trained LLMs were utilized exclusively in inference mode throughout ADEPT’s training and inference phases, without any additional finetuning.

The AdamW optimizer~\cite{adamw} was employed for its effective handling of sparse gradients and integration of weight decay with adaptive learning rates. A range of initial learning rates (1$e^{-3}$ to 1$e^{-5}$) and weight decay values (0 to 1$e^{-2}$) were explored, with the best configurations detailed in section~\ref{sec:best-combination} of the Supplementary Information. Batch sizes of 128 and 8 were used for GPT2 and Llama models, respectively, as batch size variations did not significantly affect outcomes.

Learning rates were adjusted using the ReduceLROnPlateau scheduler, with a reduction factor of 0.5, a patience period of 1 epoch, a threshold of 1$e^{-2}$, and a minimum learning rate of 1$e^{-9}$. This scheduler dynamically reduced the learning rate when model performance improvements plateaued.

\subsection{Language Modeling} \label{sec:method language modeling}

\textbf{Training:} The training of the HSM was designed to optimize the early-exit mechanism while leveraging the pre-trained strengths of the base LLM. This approach ensures that ADEPT maintains the foundational capabilities of the model while achieving an effective balance between performance and efficiency. The training process focuses on refining intermediate hidden states, guided by the early-exit criterion detailed in Section~\ref{sec:token-level early-exit}.

The HSM was trained using five early-exit thresholds (\(\delta\)) corresponding to 5\%, 10\%, 15\%, 20\%, and 25\% computational reductions (\(\mathcal{C}_{\text{threshold}}\)), optimizing it to predict final hidden states at varying depths. Cross-entropy loss was used to compare these outputs, linearly transformed into token-level distributions, with ground truth labels derived from one-position-shifted input tokens.

\textbf{Models and Datasets:}
ADEPT's performance in language modeling during the prefill phase was evaluated using two pre-trained large language models: GPT2 XL and Llama 7B. The evaluation was conducted on the Penn Treebank (PTB) \cite{penn-treebank}, WikiText103 \cite{wikitext}, and One Billion Word (1BW) \cite{one-billion-word} datasets. PTB, not included in the pre-training datasets of the base LLMs, was selected to evaluate ADEPT's ability to generalize to unseen information. In contrast, WikiText103 and 1BW, which were partially included in the pre-training data, were used to assess performance on familiar data. Additionally, ADEPT's compatibility was tested with other GPT2 variants (Large) and Llama variants (13B, 33B, and 65B) on the WikiText103 dataset to demonstrate its adaptability across different model architectures.
\begin{table*}[!t]
\caption{Benchmark results of ADEPT in language modeling compared with baselines on GPT2 XL and Llama 7B models. Lower perplexity (PTB, WikiText103, 1BW) indicates better performance.}
\label{tab:language_modeling_results}
\begin{center}
\resizebox{0.8\textwidth}{!}{%
\begin{tabular}{lccccccc}
\toprule
Method & Reduction & \multicolumn{3}{c}{GPT2 XL (Perplexity $\downarrow$)} & \multicolumn{3}{c}{Llama 7B (Perplexity $\downarrow$)} \\
\cmidrule(lr){3-5} \cmidrule(lr){6-8}
& & PTB & WikiText103 & 1BW & PTB & WikiText103 & 1BW \\
\midrule
Base Model                      & -     & 35.8 & 17.5 & 42.2 & 27.5 & 8.6  & 31.6 \\
\multirow{2}{*}{PABEE}          & 10\%  & 62.8 & 15.4 & 34.5 & 22.9 & 10.2 & 27.0 \\
                                & 20\%  & 80.5 & 17.6 & 34.8 & 24.0 & 12.1 & 26.3 \\
\multirow{2}{*}{DeeBERT}        & 10\%  & 56.9 & 17.3 & 37.8 & 24.6 & 10.1 & 28.5 \\
                                & 20\%  & 77.2 & 17.5 & 38.9 & 26.0 & 12.0 & 30.4 \\
\midrule
\multirow{2}{*}{\textbf{ADEPT (Ours)}} & 10\%  & \textbf{29.9} & \textbf{14.7} & \textbf{31.8} & \textbf{21.1} & \textbf{8.1}  & \textbf{21.8} \\
                                & 20\%  & \textbf{33.9} & \textbf{17.0} & \textbf{33.4} & \textbf{20.8} & \textbf{8.4}  & \textbf{22.7} \\
\bottomrule
\end{tabular}%
}
\end{center}
\end{table*}
\textbf{Evaluation Metrics:}
ADEPT's performance was evaluated using the optimal threshold \(\delta\), selected to meet five computational loads \(\mathcal{C}_{\text{threshold}}\) based on achieving maximum computational savings while maintaining the same perplexity as the base LLM, with the selected \(\delta\) applied consistently across all benchmarks.

The performance of the model trained with the selected threshold are presented as the representative outcomes. To further investigate the trade-off between computational efficiency and model performance, perplexities at various reduction rates were analyzed and compared. Perplexity and reduction rates were measured for single input prompts during the prefill phase, considering the maximum token capacity of the base models. The metrics were averaged across the dataset to provide a robust and comprehensive evaluation of ADEPT's effectiveness.

\textbf{Baselines:} 
For comparison, we selected early-exit methods that do not require fine-tuning base LLMs, namely PABEE~\cite{pabee} and DeeBERT~\cite{deebert}. Methods like MuE, which rely on fine-tuning, were excluded for fairness in the Language Modeling Evaluation Task. Instead, ADEPT was applied to MuE's tasks, as discussed in Section~\ref{subsec:Classfication tasks}. F-PABEE, a newer version of PABEE, was omitted due to it's unavailability for public use. 

PABEE and DeeBERT, initially designed for classification, were adapted for language modeling by replacing classifier layers with linear transformation layers. PABEE triggers early-exit when the same token is sequentially generated across layers, while DeeBERT uses entropy-based early-exit criteria. Additional transformation layers were trained on intermediate hidden states, with base LLMs kept in inference mode to avoid fine-tuning.

\begin{table}[!t]
\caption{Comparison of trainable parameters (in millions), with ratios relative to ADEPT (\(\times\)) shown in parentheses for efficiency.}
\label{tab:trainable_params}
\begin{center}
\begin{tabular}{lccc}
\toprule
\textbf{Model} & \textbf{GPT2-XL} & \textbf{Llama 7B} & \textbf{BERT} \\
\midrule
PABEE        & 3782 (\(\times\)20.5) & 4063 (\(\times\)6.7) & 258 (\(\times\)6.1) \\
DeeBERT      & 3782 (\(\times\)20.5) & 4063 (\(\times\)6.7) & 258 (\(\times\)6.1) \\
MuE          & 1558 (\(\times\)8.5)  & 6700 (\(\times\)11.1) & 345 (\(\times\)8.2) \\
\midrule
\textbf{ADEPT (Ours)} & \textbf{184} (\(\times\)1.0) & \textbf{604} (\(\times\)1.0) & \textbf{42} (\(\times\)1.0) \\
\bottomrule
\end{tabular}%
\end{center}
\end{table}

\textbf{Evaluation Results: }Table~\ref{tab:language_modeling_results} provides a comparative analysis of ADEPT against two baseline models for language modeling tasks, evaluated at 10\% and 20\% computation reduction levels across three benchmarks using GPT2 XL and Llama 7B. The first row shows the zero-shot perplexities of the unmodified pre-trained base models, which were not fine-tuned for these datasets. ADEPT consistently outperforms the baselines across all scenarios, demonstrating superior language modeling quality with fewer trainable parameters (as detailed in Table~\ref{tab:trainable_params}). Notably, ADEPT achieves a minimum 7.5\% perplexity improvement for Llama 7B on the WikiText103 benchmark and a maximum improvement of 45.0\% on the 1BW benchmark, both at a 10\% computation reduction. Comparisons with fine-tuned base LLMs are available in Section~\ref{supsec: ADEPT with base LLMs finetuned} of the Supplementary Information.

However, the baseline models show notable limitations in specific cases, such as GPT2 XL on the Penn Treebank (PTB) dataset and Llama 7B on WikiText103, highlighting their inefficiencies. Beyond enhancing language modeling quality, ADEPT delivers higher efficiency by maintaining comparable performance even at greater computation reduction rates. Further analyses and discussions on ADEPT’s advantages and trade-offs are provided in Section~\ref{ablation study}.

We attribute ADEPT's success to its token-level early-exit approach in the prefill phase, which assesses each token individually rather than relying on a single token to determine early-exit decisions for the entire prompt. This individualized approach addresses inefficiencies in traditional methods, which can lead to over-processing or under-processing of non-decision tokens, contributing to their underperformance as observed in Table~\ref{tab:language_modeling_results}. ADEPT’s flexible design can also be extended to the generation phase, leveraging its computationally efficient similarity-based metric, the lightweight Hidden State Mapper (HSM) and parallel processing of subsequent layers' Key-Values. This capability ensures both improved quality and enhanced efficiency, surpassing the benefits achieved by existing methods.

\begin{table}[!t]
\caption{Perplexity (PPL) of ADEPT when 10\% of layers are early-exited on average and Reduction (RDC) when ADEPT achieves the same perplexity as pre-trained base LLMs (GPT2 and Llama) not finetuned on WikiText103.}
\label{tab:language_modeling_results_all_models}
\begin{center}
\begin{tabular}{lcccccc}
\toprule
\textbf{Metric} & \multicolumn{2}{c}{\textbf{GPT2}} & \multicolumn{4}{c}{\textbf{Llama}} \\
\cmidrule(lr){2-3} \cmidrule(lr){4-7}
& \textbf{Large} & \textbf{XL} & \textbf{7B} & \textbf{13B} & \textbf{33B} & \textbf{65B} \\
\midrule
\textbf{PPL $\downarrow$} & 16.3 & 14.7 & 8.1 & 6.6 & 6.2 & 5.5 \\
\textbf{RDC $\uparrow$} & 29.6 & 21.4 & 21.8 & 20.5 & 17.8 & 54.9 \\
\bottomrule
\end{tabular}%
\end{center}
\end{table}

\textbf{Model Size Variants:} In Table~\ref{tab:language_modeling_results_all_models}, we evaluate ADEPT's performance with different GPT2 and Llama model variants on Wikitext103, reporting perplexity (PPL) with a 10\% reduction and the maximum achievable reduction while maintaining the baseline PPL of pre-trained, non-finetuned LLMs. ADEPT consistently demonstrates improvements similar to previous results, achieving at least a 17.8\% reduction for the base PPL (Llama 33B). These findings confirm ADEPT's scalability across models of varying sizes while preserving efficiency. Detailed data supporting these results are provided in Appendix Figure~\ref{fig:lm-all} and Table~\ref{tab: language modeling details}.

\subsection{Classification Tasks}
\label{subsec:Classfication tasks}
\textbf{Training:} The training process consisted of two stages. First, a BERT model \cite{bert} was fine-tuned for the target tasks without incorporating ADEPT. Concurrently, a pre-trained OFA model \cite{wang2022ofa} was adapted for the SNLI-VE (Visual Entailment) task \cite{SNLI-VE2019}. In the second stage, ADEPT was integrated with these fine-tuned models, and the Hidden State Mapper (HSM) was trained while keeping the base models fixed. The training objective for ADEPT was to achieve a specified reduction in computational load, guided by the early-exit metric.

The HSM predicts anticipated final hidden states for tokens that qualify for early exit, which are directly utilized by the classifier layer. Unlike language modeling, classification tasks lack token-level transformation, preventing the use of one-position-shifted input tokens as labels. Instead, similarity loss is employed, using the original final hidden states as labels to train the HSM to generate representations closely aligned with the original outputs. Additionally, a task-specific binary cross-entropy loss is combined with the similarity loss to form the total loss, ensuring robust training and task performance.

\textbf{Models and Datasets:}  
ADEPT's performance in textual classification tasks was evaluated using four single-label classification (SLC) tasks from the GLUE Benchmark \cite{glue} with the BERT model \cite{bert}. These tasks were chosen to ensure a direct and fair comparison with existing methods, including DeeBERT, PABEE, and F-PABEE \cite{deebert,pabee,fpabee}. For multimodal classification, ADEPT was tested on the SNLI-VE (Visual Entailment) dataset \cite{SNLI-VE2019}, leveraging the OFA model \cite{wang2022ofa}, which combines an encoder-decoder architecture. This multimodal setup also facilitates comparison with MuE \cite{mue}, a method designed for similar tasks.

\textbf{Evaluation Metrics:} 
ADEPT's performance in classification tasks is evaluated using classification accuracy and processing acceleration. Baseline comparisons were drawn from performance metrics reported in the literature for comparable acceleration levels \cite{fpabee,deebert,mue}. This focus on efficiency and accuracy provides a comprehensive assessment of ADEPT's effectiveness relative to other models.

\begin{table*}[!t]
\caption{Comparison of classification benchmarking results. Results of existing works are from F-PABEE~\cite{fpabee} and DeeBERT~\cite{deebert}. Metrics include accuracy (ACC) $\uparrow$ and reduction in computation (RDC) $\downarrow$ on RTE, SST-2, MNLI, and QQP datasets.}
\label{tab:classification_task_results}
\begin{center}
\resizebox{\textwidth}{!}{%
\begin{tabular}{lcccccccccc}
\toprule
\textbf{Method} & \textbf{Params}$\downarrow$ & \multicolumn{4}{c}{\textbf{RTE}} & \multicolumn{4}{c}{\textbf{SST-2}} \\
\cmidrule(lr){3-6} \cmidrule(lr){7-10}
& \textbf{[M]} & \textbf{ACC [\%]} & \textbf{RDC [\%]} & \textbf{ACC [\%]} & \textbf{RDC [\%]} & \textbf{ACC [\%]} & \textbf{RDC [\%]} & \textbf{ACC [\%]} & \textbf{RDC [\%]} \\
\midrule
BERT base      & 345 & 69.1 & 0   & 69.1 & 0   & 91.3 & 0   & 91.3 & 0   \\ \midrule
BranchyNet     & 258 & 54.7 & -76 & 67.4 & -47 & 79.9 & -76 & 88.3 & -49 \\
Shallow-Deep   & 258 & 54.7 & -76 & 67.2 & -48 & 79.5 & -77 & 88.4 & -48 \\
PABEE          & 258 & 55.8 & -75 & 67.7 & -46 & 79.9 & -77 & 88.7 & -48 \\
F-PABEE        & 258 & 56.0 & -76 & 68.1 & -47 & 80.5 & -76 & \textbf{92.3} & -48 \\
DistilBERT     & 258 &  -   & -75 & 59.7 & -40 &   -  & -75 & 89.9 & -40 \\
DeeBERT        & 258 &  -   & -75 & 65.9 & -33 &   -  & -75 & 89.2 & -47 \\
\midrule
\textbf{Ours}  & \textbf{42}  & \textbf{60.8} & -75 & \textbf{68.8} & -53 & \textbf{85.8} & -75 & \underline{90.3} & -49 \\
\midrule
\textbf{Method} & \textbf{Params}$\downarrow$ & \multicolumn{4}{c}{\textbf{MNLI}} & \multicolumn{4}{c}{\textbf{QQP}} \\
\cmidrule(lr){3-6} \cmidrule(lr){7-10}
& \textbf{[M]} & \textbf{ACC [\%]} & \textbf{RDC [\%]} & \textbf{ACC [\%]} & \textbf{RDC [\%]} & \textbf{ACC [\%]} & \textbf{RDC [\%]} & \textbf{ACC [\%]} & \textbf{RDC [\%]} \\
\midrule
BERT base      & 345 & 83.1 & 0   & 83.1 & 0   & 89.2 & 0   & 89.2 & 0   \\ \midrule
BranchyNet     & 258 & 63.8 & -76 & 78.3 & -53 & 71.6 & -80 & 89.3 & -50 \\
Shallow-Deep   & 258 & 64.1 & -77 & 78.2 & -51 & 71.4 & -79 & 89.6 & -51 \\
PABEE          & 258 & 63.9 & -77 & 78.9 & -52 & 68.6 & -82 & 89.6 & -49 \\
F-PABEE        & 258 & 66.9 & -72 & \textbf{83.9} & -53 & \textbf{79.6} & -82 & \textbf{90.8} & -49 \\
DistilBERT     & 258 &  -   & -75 & 78.6 & -40 &   -  & -75 & 88.1 & -40 \\
DeeBERT        & 258 &  -   & -75 & 79.2 & -37 &   -  & -75 & 87.2 & -49 \\
\midrule
\textbf{Ours}  & \textbf{42}  & \textbf{71.8} & -75 & \underline{80.5} & -52 & \underline{77.8} & -75 & 88.6 & -50 \\
\bottomrule
\end{tabular}%
}
\end{center}
\end{table*}

\begin{table}[!t]
\caption{Multimodal benchmarking results on SNLI-VE using the OFA Base model. Results of existing methods are taken from the MuE paper~\cite{mue}. Metrics include accuracy (ACC) and computation reduction in computation (RDC).}
\label{tab:ofa_results}
\begin{center}
\begin{tabular}{lccc}
\toprule
\textbf{Method} & \textbf{Params [M]} $\downarrow$ & \textbf{ACC [\%]} $\uparrow$ & \textbf{RDC [\%]}  $\downarrow$\\
\midrule
OFA Base        & 182 & 89.2 & 0.0  \\ \midrule
DeeBERT         & 548 & 78.8 & -15.0 \\
PABEE           & 548 & 85.2 & -15.3 \\
MuE             & 182 & \textbf{88.5} & -50.0 \\
\midrule
\multirow{2}{*}{\textbf{Ours}}   & \multirow{2}{*}{\textbf{85}} & \textbf{86.2} & -15.0 \\
                &              & \underline{83.3} & -48.0 \\
\bottomrule
\end{tabular}%
\end{center}
\end{table}

\textbf{Evaluation Results:}  
Table~\ref{tab:classification_task_results} summarizes the classification accuracy of the BERT model with ADEPT across various tasks, demonstrating 50\% and 75\% latency reductions, corresponding to 2× and 4× acceleration, respectively. Despite utilizing only one-sixth of the parameters compared to existing methods, ADEPT outperforms them on RTE, SST-2, and MNLI at a 75\% reduction rate and delivers comparable performance in other scenarios.

Table~\ref{tab:ofa_results} highlights the performance of the OFA model with ADEPT on the SNLI-VE task, achieving 15\% and 48\% time reductions. ADEPT surpasses DeeBERT and PABEE at a 15\% reduction rate and retains strong performance even at a 50\% reduction. While ADEPT falls short of MuE at higher reduction levels, this can be attributed to MuE's reliance on fine-tuning the entire OFA model, which involves 182M parameters, double the size of ADEPT.
\section{Discussion}
\textbf{Addition Module for Early Exit Decision:} The similarity-based early-exit metric offers substantial computational efficiency compared to module-based approaches. For instance, in GPT-2 XL, module-based metrics demand 161 MFLOPs per token per layer, surpassing the 119 MFLOPs needed for regular processing, making early exits impractical unless they occur very early (e.g., before the 20th layer in a 48-layer model). By utilizing a lightweight convergence similarity metric, ADEPT avoids additional module computations, reducing overhead while maintaining task performance.

\textbf{Additional Computational Requirement for KV of Skipped Layers:} In models like GPT2 XL and Llama 7B, the additional computational cost of generating Key-Values (KV) is minor, contributing only 8.5\% and 9.5\% of a layer's computation, respectively. For instance, in GPT2 XL, exiting all tokens at the 38th of 48 layers reduces computation by 18.1\%, increasing slightly to 19.7\% when KV generation is omitted. This 1.6\% difference is negligible, considering KV generation ensures multi-head attention (MHA) consistency for tokens that do not exit early.

\textbf{Trainable Parameters:} The number of trainable parameters is comprehensively outlined in Table~\ref{tab:trainable_params}. ADEPT has an order of magnitude lower trainable parameters than others, meaning lower training cost. Further details about the HSM and the number of trainable parameters of other early-exit methods are shown in Appendix Sections ~\ref{supsec:HSM details} and ~\ref{sec: trainable parameter calculation explanation}.

\textbf{Ablation Studies:}  
We conduct extensive ablation studies detailed in the Appendix. These include: ADEPT's performance with fine-tuned base models (Section~\ref{supsec: ADEPT with base LLMs finetuned}); evaluation across all GPT2 and Llama1 variants (Section~\ref{supsec:ADEPT on All GPT2 and Llama Models}); analysis of early-exit distributions under different thresholds (Section~\ref{supsec:Distribution of Early-Exit with Different Threshold Values}) and across various methods (Section~\ref{supsec:Comparison of Early-Exit Distribution}); training with varying thresholds (Section~\ref{supsec:Training on Different Thresholds}); comparison of prompt-level (all-at-once) vs. token-level early exits (Section~\ref{supsec:All-at-Once Early-Exit}); and the impact of local vs. global similarity metrics with inertia (Section~\ref{supsec:Local vs Global Metric}).

\section{Conclusion}
In this work, we introduced ADEPT, a novel framework that enables efficient token-level early exits in Transformer models for both prefill and generation phases. Unlike prior methods that rely on prompt-level or single-token exits, ADEPT leverages a Markov Decision Process with preference-based rewards to dynamically optimize computational efficiency while maintaining task-specific performance. By incorporating an adaptive mechanism and optimizing key-value (KV) generation for skipped layers, ADEPT effectively reduces computational overhead while preserving model quality.  

Extensive experimental evaluations demonstrate the effectiveness of our approach, achieving up to 25\% computational savings in language modeling tasks, 4× speed-ups in classification, and up to 45\% performance enhancements. ADEPT’s lightweight and flexible design makes it adaptable to various pre-trained models, ensuring broad applicability across Transformer-based architectures. Moreover, by systematically managing sequential dependencies in KV caching, ADEPT enables more reliable and efficient early exits without sacrificing model coherence.  

\section{Impact Statement}
ADEPT effectively balances efficiency and performance by utilizing a threshold-based early-exit mechanism. In this work, thresholds were determined heuristically, exploring various optimization strategies. However, the trade-off between performance and efficiency may vary across input domains, potentially leading to scenarios where early-exit is not achieved. This limitation is inherent to all early-exit methods that rely on heuristically chosen metrics, such as PABEE, F-PABEE, DeeBERT, and MuE. Despite this, ADEPT's token-by-token early-exit approach during the prefill phase offers a greater likelihood of achieving early exits compared to prompt-level methods used in existing techniques.

To further enhance the early-exit methods, including ADEPT, future work should focus on developing algorithms capable of dynamically adjusting thresholds based on input characteristics. This would ensure consistent performance and efficiency across diverse tasks and input domains.
\bibliographystyle{unsrt}

\newpage
\appendix

\setcounter{equation}{0}
\setcounter{table}{0}
\setcounter{figure}{0}
\renewcommand*{\theequation}{{S}\arabic{equation}}
\renewcommand*{\theHequation}{{S}\arabic{equation}}
\renewcommand*{\thetable}{{S}\arabic{table}}
\renewcommand*{\theHtable}{{S}\arabic{table}}
\renewcommand*{\thefigure}{{S}\arabic{figure}}
\renewcommand*{\theHfigure}{{S}\arabic{figure}}

\section{Proof of Preference-Based Reward Formulation:}
\label{supsec:proof_formulation}
The preference probability \(\mathcal{P}_\pi(a_{i,j} | s_{i,j})\) is constructed to represent the most unbiased probability distribution given known constraints. Following the principle of maximum entropy, we model the probability of actions \(a_{i,j}\) at state \(s_{i,j}\) such that no additional assumptions are introduced beyond the constraints imposed by task-specific performance \(\mathcal{T}_{i,j}\) and computational cost \(\mathcal{C}_{i,j}\).

The entropy of the distribution is defined as:
\[
H(\mathcal{P}_\pi) = -\sum_{a \in A} \mathcal{P}_\pi(a | s_{i,j}) \log \mathcal{P}_\pi(a | s_{i,j}),
\]
where \(A = \{\text{Continue}, \text{Exit}\}\) represents the action space.

\paragraph{Constraints on the Probability Distribution:}
To incorporate the trade-off between task-specific performance and computational cost, we define the expected reward constraints as:
\[
\mathbb{E}_{\mathcal{P}_\pi}\left[\mathcal{T}_{i,j}\right] = \sum_{a \in A} \mathcal{P}_\pi(a | s_{i,j}) \mathcal{T}_{i,j}(a),
\]
\[
\mathbb{E}_{\mathcal{P}_\pi}\left[\mathcal{C}_{i,j}\right] = \sum_{a \in A} \mathcal{P}_\pi(a | s_{i,j}) \mathcal{C}_{i,j}(a),
\]
where \(\mathcal{T}_{i,j}(a)\) and \(\mathcal{C}_{i,j}(a)\) represent the task-specific performance and computational cost associated with action \(a\), respectively.

\paragraph{Maximum Entropy Distribution:}
To derive the distribution \(\mathcal{P}_\pi(a | s_{i,j})\) that maximizes entropy while respecting the given constraints, we maximize the entropy function:
\[
H(\mathcal{P}_\pi) = -\sum_{a \in A} \mathcal{P}_\pi(a | s_{i,j}) \log \mathcal{P}_\pi(a | s_{i,j}),
\]
subject to the following constraints:
1. The probability distribution sums to 1:
\[
\sum_{a \in A} \mathcal{P}_\pi(a | s_{i,j}) = 1.
\]
2. The expected task-specific performance, \(\mathcal{T}_{i,j}(a)\), is fixed:
\[
\mathbb{E}_{\mathcal{P}_\pi}\left[\mathcal{T}_{i,j}\right] = \sum_{a \in A} \mathcal{P}_\pi(a | s_{i,j}) \mathcal{T}_{i,j}(a).
\]
3. The expected computational cost, \(\mathcal{C}_{i,j}(a)\), is fixed:
\[
\mathbb{E}_{\mathcal{P}_\pi}\left[\mathcal{C}_{i,j}\right] = \sum_{a \in A} \mathcal{P}_\pi(a | s_{i,j}) \mathcal{C}_{i,j}(a).
\]

\textbf{Using Lagrange Multipliers:} 
To solve this optimization problem, we construct the Lagrangian:
\[
\mathcal{L}(\mathcal{P}_\pi, \lambda, \beta, \gamma) = -\sum_{a \in A} \mathcal{P}_\pi(a | s_{i,j}) \log \mathcal{P}_\pi(a | s_{i,j})
+ \lambda \left(\sum_{a \in A} \mathcal{P}_\pi(a | s_{i,j}) \mathcal{T}_{i,j}(a) - \mathbb{E}_{\mathcal{P}_\pi}[\mathcal{T}_{i,j}]\right)
\]
\[
+ \beta \left(\sum_{a \in A} \mathcal{P}_\pi(a | s_{i,j}) \mathcal{C}_{i,j}(a) - \mathbb{E}_{\mathcal{P}_\pi}[\mathcal{C}_{i,j}]\right)
+ \gamma \left(\sum_{a \in A} \mathcal{P}_\pi(a | s_{i,j}) - 1\right),
\]
where \(\lambda\), \(\beta\), and \(\gamma\) are the Lagrange multipliers associated with the task-specific performance, computational cost, and probability normalization constraints, respectively.

\textbf{First-Order Condition:}
To maximize \(\mathcal{L}\), we take the derivative with respect to \(\mathcal{P}_\pi(a | s_{i,j})\) and set it to zero:
\[
\frac{\partial \mathcal{L}}{\partial \mathcal{P}_\pi(a | s_{i,j})} = -\left(1 + \log \mathcal{P}_\pi(a | s_{i,j})\right) + \lambda \mathcal{T}_{i,j}(a) - \beta \mathcal{C}_{i,j}(a) + \gamma = 0.
\]
Rearranging terms gives:
\[
\log \mathcal{P}_\pi(a | s_{i,j}) = -1 + \gamma + \lambda \mathcal{T}_{i,j}(a) - \beta \mathcal{C}_{i,j}(a).
\]

Exponentiating both sides yields:
\[
\mathcal{P}_\pi(a | s_{i,j}) = \exp\left(-1 + \gamma + \lambda \mathcal{T}_{i,j}(a) - \beta \mathcal{C}_{i,j}(a)\right).
\]

\textbf{Simplification Using Normalization:}

The normalization constraint \(\sum_{a \in A} \mathcal{P}_\pi(a | s_{i,j}) = 1\) allows us to determine the term \(\exp(-1 + \gamma)\), leading to:
\[
\mathcal{P}_\pi(a | s_{i,j}) = \frac{\exp\left(\lambda \mathcal{T}_{i,j}(a) - \beta \mathcal{C}_{i,j}(a)\right)}{\sum_{a' \in A} \exp\left(\lambda \mathcal{T}_{i,j}(a') - \beta \mathcal{C}_{i,j}(a')\right)}.
\]

For simplicity, this is often expressed in proportional form:
\[
\mathcal{P}_\pi(a | s_{i,j}) \propto \exp\left(\lambda \mathcal{T}_{i,j}(a) - \beta \mathcal{C}_{i,j}(a)\right),
\]
where \(\lambda > 0\) and \(\beta > 0\) represent the relative importance of task-specific performance and computational cost, respectively.

\paragraph{Specific Case for Action \(a_{i,j}\):}
For a specific action \(a_{i,j}\), substituting the task-specific performance \(\mathcal{T}_{i,j}\) and computational cost \(\mathcal{C}_{i,j}\), the parameterized preference probability is:
\[
\mathcal{P}_\pi(a_{i,j} | s_{i,j}) \propto \exp\left(\lambda \cdot \mathcal{T}_{i,j}(a_{i,j}) - \beta \cdot \mathcal{C}_{i,j}(a_{i,j})\right).
\]

\paragraph{Log-Likelihood Reward Function:}
The reward function is defined as the log-likelihood of the preference probability to ensure alignment with policy gradient methods:
\[
R(s_{i,j}, a_{i,j}) = \log \mathcal{P}_\pi(a_{i,j} | s_{i,j}),
\]
which simplifies to:
\[
R(s_{i,j}, a_{i,j}) = \lambda \cdot \mathcal{T}_{i,j}(a_{i,j}) - \beta \cdot \mathcal{C}_{i,j}(a_{i,j}) + \text{constant}.
\]

\paragraph{Interpretation:}
This formulation ensures the agent learns to prioritize actions balancing high performance and low computational cost. The parameters \(\lambda\) and \(\beta\) allow fine-tuning of this trade-off, aligning the reward structure with the model's optimization goals.

\section{HSM Module Architecture} \label{supsec:HSM details}

\begin{figure}[ht]
\begin{center}
\centerline{\includegraphics[width=0.55\columnwidth]{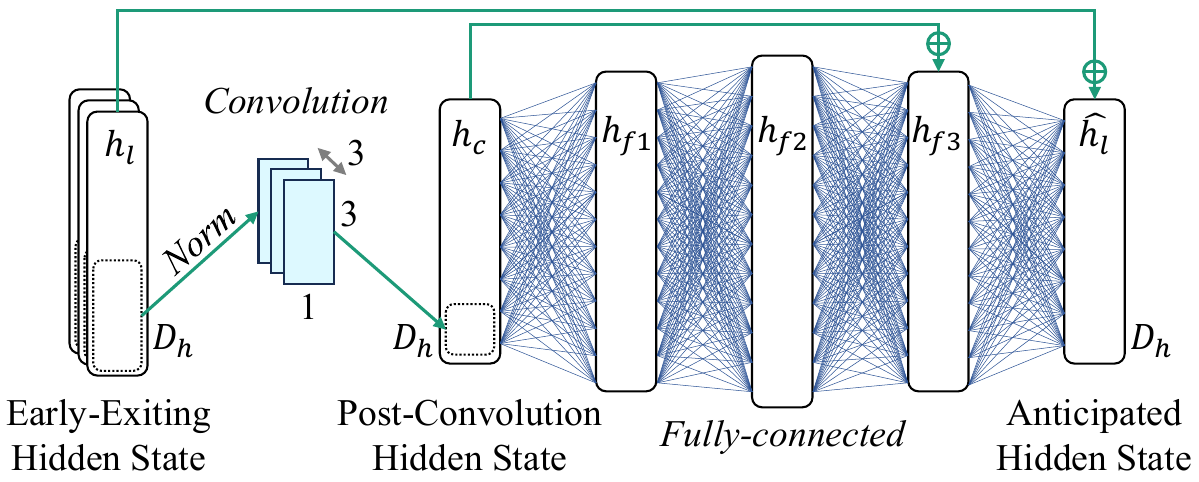}}
\vspace{-2mm}
\caption{Architecture of Hidden State Mapper. It takes neighboring hidden states ($h_{l}$ and $h_{l-1}$) and history states ($h_{intertia}$) as inputs and outputs the anticipated final hidden state ($\hat{h}_{l}$).}
\label{fig:hsm}
\end{center}
\vspace{-8mm}
\end{figure}

\begin{table}[ht]
\caption{Architecture details of the ADEPT module. Dimensions are denoted as \(D_h\) (hidden dimension).}
\label{Supple:HSM architecture details}

\begin{center}
\resizebox{0.5\columnwidth}{!}{%
\begin{tabular}{lcccc}
\toprule
\textbf{Layer}          & \textbf{Dim$_\text{Input}$} & \textbf{Dim$_\text{Output}$} & \textbf{Stride} & \textbf{Pad} \\
\midrule
Normalization           & \(D_h\)                     & \(D_h\)                      & -               & -            \\
Convolution             & 3, \(D_h\)                  & 1, \(D_h\)                   & 1, 1           & 0, 1         \\
\midrule
\multicolumn{5}{l}{\textbf{MLP:}} \\
Fully-connected 1       & \(D_h\)                     & \(4D_h\)                     & -               & -            \\
Fully-connected 2       & \(4D_h\)                    & \(8D_h\)                     & -               & -            \\
Fully-connected 3       & \(8D_h\)                    & \(4D_h\)                     & -               & -            \\
Normalization           & \(4D_h\)                    & \(4D_h\)                     & -               & -            \\
Add                     & \(4D_h\)                    & \(4D_h\)                     & -               & -            \\
Fully-connected 4       & \(4D_h\)                    & \(D_h\)                      & -               & -            \\
\midrule
Normalization           & \(D_h\)                     & \(D_h\)                      & -               & -            \\
Add                     & \(D_h\)                     & \(D_h\)                      & -               & -            \\
\bottomrule
\end{tabular}%
}
\end{center}
\vspace{-3mm}
\end{table}

Table~\ref{Supple:HSM architecture details} lists the details of the ADEPT module and Figure~\ref{fig:hsm} illustrates it. The input of the module includes the early-exiting hidden state, the hidden state from the previous layer, and the history state that was discussed in Section~\ref{sec:token-level early-exit}. The three states are separately normalized by the first RMS normalization layer, and the normalized states are formed in the shape of (3, hidden state dimension). The following convolution layer takes it and generates the outputs with the same dimension size as the hidden state. 4 fully-connected layers and normalization layers are followed as displayed in Table~\ref{Supple:HSM architecture details}.

\section{Calculation of Trainable Parameters of Early-Exit Mehtods} \label{sec: trainable parameter calculation explanation}
We detail the calculation of trainable parameters for existing early-exit methods and ADEPT, as presented in Table~\ref{tab:trainable_params}. The parameter values for PABEE and DEEBERT in Table~\ref{tab:trainable_params} exclude the base model parameters. For GPT2 XL, which consists of 48 decoder layers, these methods introduce an additional linear transformation layer for each decoder layer, excluding the final layer that already includes this layer, resulting in 47 additional layers. As outlined in Section~\ref{sec:method language modeling}, GPT2 XL has a hidden dimension size of 1600 and a vocabulary size of 50257. Each linear transformation layer requires a weight matrix of size \(1600 \times 50257\) and a bias vector of size 50257. The total parameters for these layers are computed as \(47 \times (1600 \times 50257 + 50257)\), amounting to 3782M. 

For MuE, the parameter count matches the base model since it requires fine-tuning the entire model. In contrast, ADEPT achieves a significantly reduced parameter count by sharing a single Hidden State Mapper (HSM) across all decoder layers. For GPT2 XL, this design reduces the additional trainable parameters to 184M, highlighting the efficiency of the ADEPT framework.

\section{Training Optimizer Parameters} \label{sec:best-combination}

We employed the AdamW optimizer~\cite{adamw} with varying hyperparameters, including the initial learning rate and weight decay, across different experiments.

For GPT2-based experiments with ADEPT, DeeBERT, and PABEE on PTB, Wikitext103, and 1BW, we used an initial learning rate of \(1 \times 10^{-4}\) and a weight decay of 0.

For Llama-based experiments, the hyperparameters varied by task:
\begin{enumerate}
\item \textbf{ADEPT}: An initial learning rate of \(2 \times 10^{-5}\), \(5 \times 10^{-5}\), and \(5 \times 10^{-5}\) with weight decays of \(1 \times 10^{-2}\), 0, and 0 were used for PTB, Wikitext103, and 1BW, respectively.

\item \textbf{DeeBERT and PABEE}: Initial learning rates of \(3 \times 10^{-5}\), \(5 \times 10^{-5}\), and \(4 \times 10^{-5}\) with weight decays of \(1 \times 10^{-1}\), 0, and 0 were used for PTB, Wikitext103, and 1BW, respectively.
\end{enumerate}

For BERT-based experiments on RTE, SST-2, MNLI, and QQP:
\begin{enumerate}
\item Initial learning rates were \(1 \times 10^{-5}\), \(1 \times 10^{-3}\), \(1 \times 10^{-5}\), and \(2 \times 10^{-5}\), respectively.

\item Corresponding weight decays were \(1 \times 10^{-1}\), 0, 0, and 0.
\end{enumerate}

For OFA-based experiments, we used an initial learning rate of \(2 \times 10^{-4}\) and a weight decay of 0.

\section{ADEPT with Base LLMs Finetuned} \label{supsec: ADEPT with base LLMs finetuned}

Additional experiments to study performance improvement by ADEPT on fine-tuned LLMs based on GPT2 X-Large model and the Penn Treebank (PTB) dataset are presented. We summarized the results in Table~\ref{tab: Fine-tuned ADEPT}. ADEPT reduces computation by 3\% while maintaining the performance of the fine-tuned model without early-exit methods and allows the model to perform better (6\% improvement in perplexity) with 1\% computation reduction. The improvement in performance and computation efficiency are not as high as ADEPT with pre-trained LLMs when ADEPT is applied to the fine-tuned LLMs. However, considering ADEPT is originally designed to work with pre-trained LLMs, the performance-efficiency optimization can be further improved if ADEPT with fine-tuning is re-designed in the future.

\begin{table}[!t]
\caption{Performance comparison of fine-tuned LLM with and without ADEPT on GPT2 XL using the Penn Treebank dataset. Metrics include reduction in computation (\%) and perplexity (lower is better).}
\label{tab: Fine-tuned ADEPT}

\begin{center}
\resizebox{0.6\columnwidth}{!}{%
\begin{tabular}{lcc}
\toprule
\textbf{Method} & \textbf{Reduction [\%]} $\uparrow$ & \textbf{Perplexity} $\downarrow$ \\
\midrule
Fine-tuned LLM without Early-Exit & -  & 13.9 \\
\multirow{6}{*}{Fine-tuned LLM with ADEPT} 
                                  & 1  & 13.1 \\
                                  & 3  & 13.9 \\
                                  & 5  & 14.6 \\
                                  & 10 & 16.0 \\
                                  & 15 & 17.6 \\
                                  & 20 & 19.3 \\
\bottomrule
\end{tabular}%
}
\end{center}
\vspace{-3mm}
\end{table}

\section{Performance on Language Modeling}
\begin{table*}[!ht]
\caption{Performance comparison of ADEPT and baseline methods in language modeling across GPT2 XL and Llama 7B on PTB, WikiText103, and 1BW datasets. Metrics include perplexity (lower is better) and reduction in computation (\%).}
\label{tab: language modeling details}

\begin{center}
\resizebox{0.8\textwidth}{!}{%
\begin{tabular}{lcccccccc}
\toprule
\textbf{Method} & \textbf{Reduction [\%]} & \multicolumn{3}{c}{\textbf{GPT2 XL}} & \multicolumn{3}{c}{\textbf{Llama 7B}} \\
\cmidrule(lr){3-5} \cmidrule(lr){6-8}
& & \textbf{PTB} & \textbf{WikiText103} & \textbf{1BW} & \textbf{PTB} & \textbf{WikiText103} & \textbf{1BW} \\
\midrule
\textbf{Baseline}          & 0   & 35.8 & 17.5 & 42.2 & 27.5 & 8.6 & 31.6 \\
\midrule
\multirow{6}{*}{PABEE}     & 1   & 40.5 & 15.2 & 37.5 & 23.4 & 8.8 & 27.7 \\
                           & 5   & 51.6 & 15.2 & 34.6 & 22.8 & 9.5 & 27.4 \\
                           & 10  & 62.8 & 15.4 & 34.5 & 22.9 & 10.2 & 27.0 \\
                           & 15  & 68.5 & 15.8 & 34.4 & 23.3 & 11.2 & 26.4 \\
                           & 20  & 80.5 & 17.6 & 34.8 & 24.0 & 12.1 & 26.3 \\
                           & 25  & 92.1 & 17.8 & 35.8 & 26.5 & 13.0 & 27.4 \\
\midrule
\multirow{6}{*}{DeeBERT}   & 1   & 35.3 & 17.3 & 36.9 & 23.4 & 8.9 & 27.4 \\
                           & 5   & 44.8 & 17.3 & 37.3 & 23.6 & 9.3 & 27.8 \\
                           & 10  & 56.9 & 17.3 & 37.8 & 24.6 & 10.1 & 28.5 \\
                           & 15  & 65.1 & 17.5 & 38.3 & 25.3 & 11.0 & 29.4 \\
                           & 20  & 77.2 & 17.5 & 38.9 & 26.0 & 12.0 & 30.4 \\
                           & 25  & 86.8 & 17.7 & 39.7 & 26.8 & 13.0 & 31.2 \\
\midrule
\multirow{6}{*}{\textbf{ADEPT (Ours)}} 
                           & \textbf{1}   & \textbf{25.9} & \textbf{13.4} & \textbf{31.6} & \textbf{18.6} & \textbf{7.9} & \textbf{20.6} \\
                           & \textbf{5}   & \textbf{27.9} & \textbf{13.9} & \textbf{31.7} & \textbf{20.4} & \textbf{8.0} & \textbf{21.7} \\
                           & \textbf{10}  & \textbf{29.9} & \textbf{14.7} & \textbf{31.8} & \textbf{21.1} & \textbf{8.1} & \textbf{21.8} \\
                           & \textbf{15}  & \textbf{31.8} & \textbf{15.8} & \textbf{32.4} & \textbf{20.6} & \textbf{8.2} & \textbf{21.9} \\
                           & \textbf{20}  & \textbf{33.9} & \textbf{17.0} & \textbf{33.4} & \textbf{20.8} & \textbf{8.4} & \textbf{22.7} \\
                           & \textbf{25}  & \textbf{36.2} & \textbf{18.6} & \textbf{35.0} & \textbf{23.1} & \textbf{8.8} & \textbf{26.1} \\
\bottomrule
\end{tabular}%
}
\end{center}
\vspace{-3mm}
\end{table*}

\begin{figure}[!ht]
\begin{center}
\centerline{\includegraphics[scale=0.45]{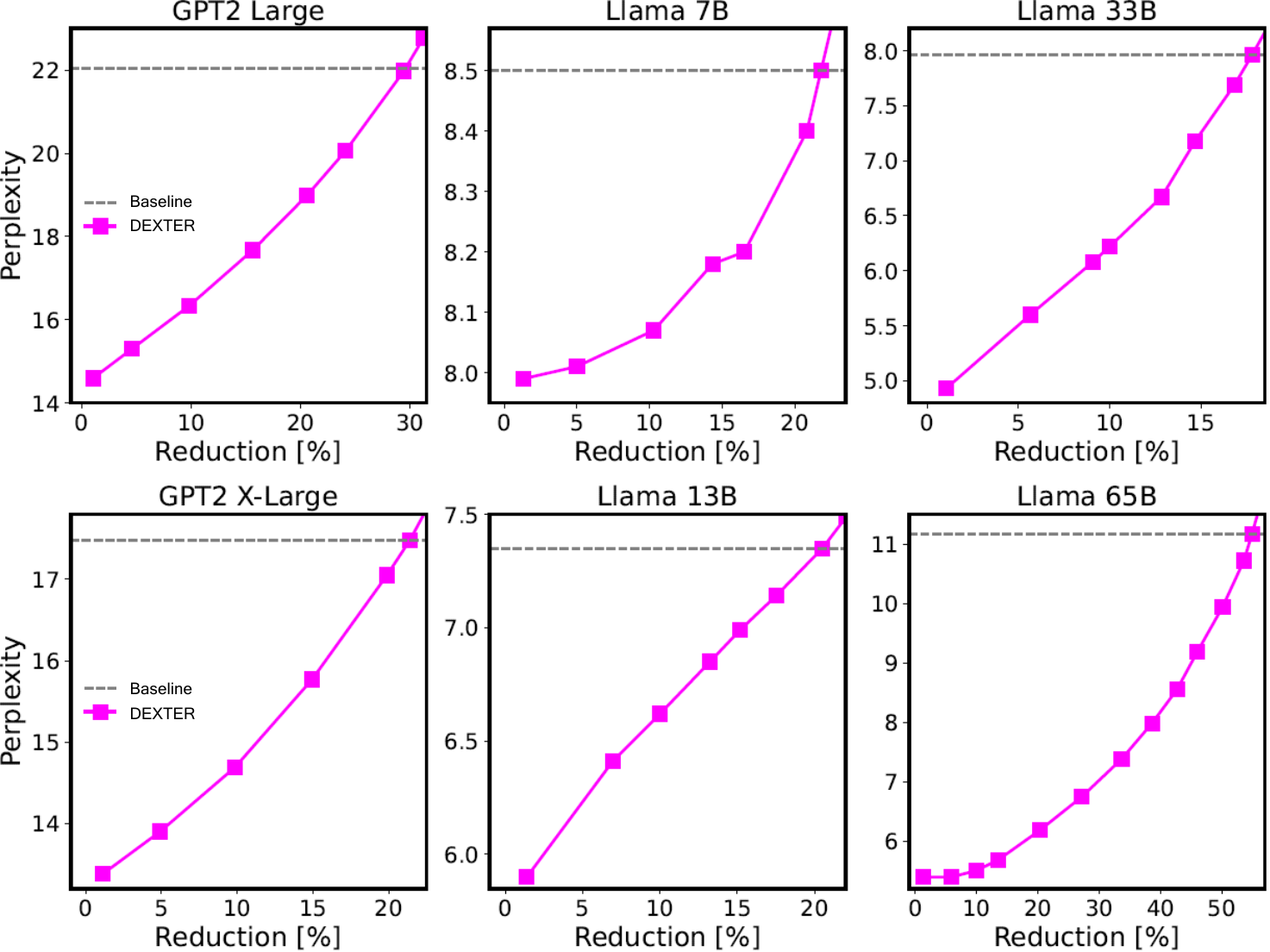}}
\caption{Performance of ADEPT with GPT2 Large model and Llama 13B, 33B and 65B on WikiText103.}
\label{fig:lm-all}
\end{center}
\end{figure}

\subsection{Comparison with Baselines}

Table~\ref{tab: language modeling details} presents extended data about the performance of ADEPT on GPT2 XL and Llama 7B models for PTB, Wikitext103, and 1BW. Numbers on the 3rd to the 8th columns represent perplexities where the methods reduce computation as much as numbers on the reduction column.

\subsection{ADEPT on All GPT2 and Llama Models}~\label{supsec:ADEPT on All GPT2 and Llama Models}
Figure~\ref{fig:lm-all} illustrates the performance of ADEPT on GPT2 L and XL model, and Llama 7B, 13B, 33B, and 65B for Wikitext103 dataset. The dotted line represents the performance of the base model. As displayed in the figure, reduction and perplexity (performance) are in the opposite relationship. 

\subsection{Distribution of Early-Exit with Different Threshold Values}~\label{supsec:Distribution of Early-Exit with Different Threshold Values}

\begin{figure}[!ht]
\begin{center}
\centerline{\includegraphics[scale=0.5]{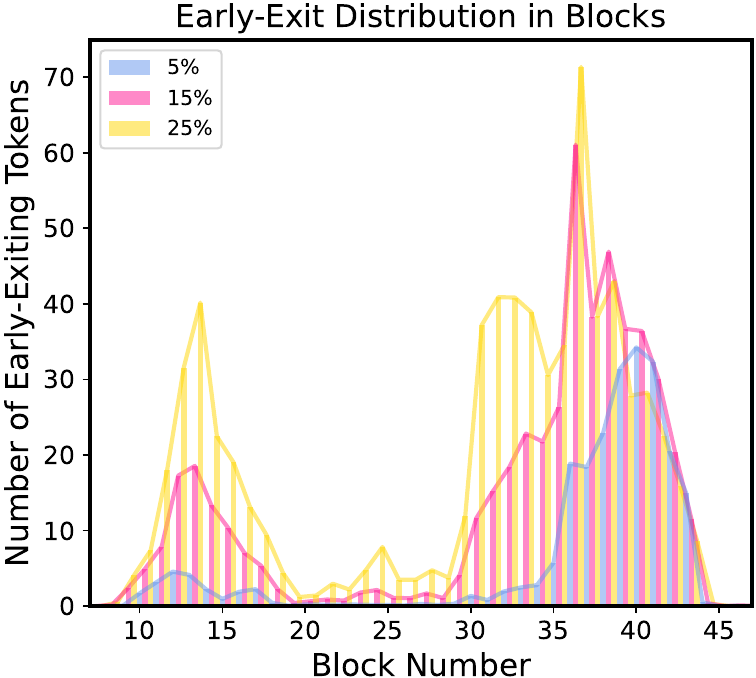}}
\caption{Distribution of early-exit per decoder layer of GPT2 X-Large model with ADEPT on WikiText103 benchmark for an input prompt in the prefill phase with the maximum length. Tokens independently early-exit the layer chain depending on the similarities and the threshold.}
\label{early-exit distribution in blocks}
\end{center}
\end{figure}

Figure~\ref{early-exit distribution in blocks} shows the number of early-exiting tokens per layer under three different early-exiting thresholds each of which induces the reduction of 5\%, 15\%, and 25\% on average, respectively. When the threshold decreases for higher reduction, the distribution of early-exiting tokens becomes wide. It double-confirms that the token-by-token early-exiting logic of ADEPT allows to dynamically early-exit independently, leading to improved efficiency while minimizing performance degradation.\par

\section{Comparison of Early-Exit Distribution}
\label{supsec:Comparison of Early-Exit Distribution}

\begin{figure}[ht]
\begin{center}
\centerline{\includegraphics[width=0.45\columnwidth]{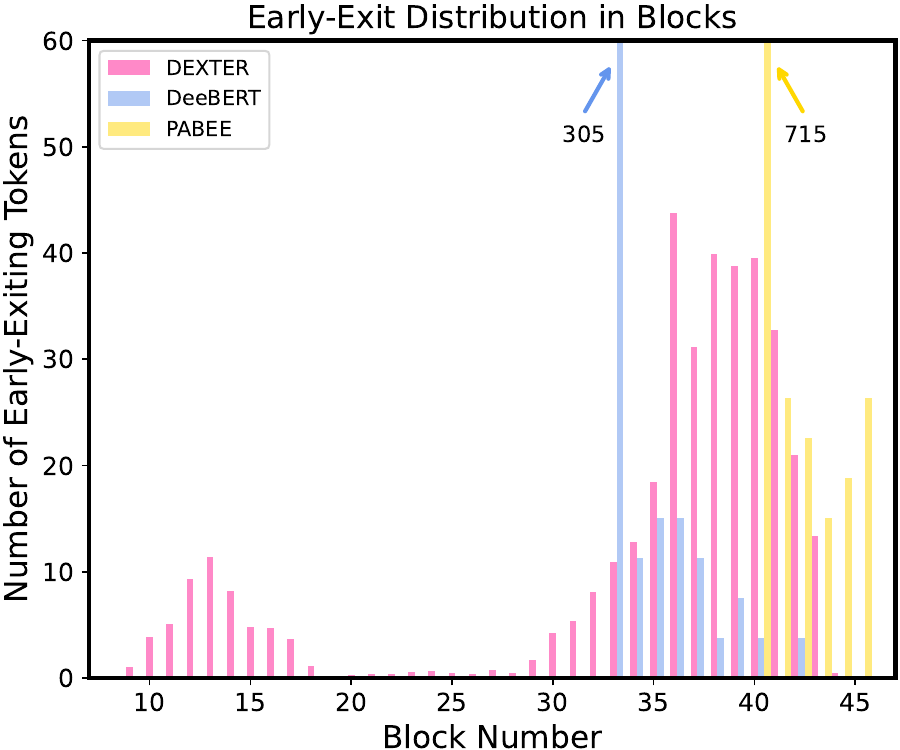}}
\caption{Distribution of early-exit per decoder layer of GPT2 X-Large model on WikiText103 benchmark for a prompted sentence with the maximum length. Tokens independently early-exit the decoder layers depending on the similarities and the threshold.}
\label{early-exit distribution in blocks2}
\end{center}
\end{figure}

The promising performance of ADEPT is attributed to token-by-token independent and dynamic early-exiting adaptive to the individual token's difficulty. Figure~\ref{early-exit distribution in blocks2} shows the average of the distribution of early-exiting tokens along layers of ADEPT and 2 baselines. In ADEPT, the early-exiting locations in layers are more distributed than the others thanks to early-exiting on the token-by-token basis. Mathematically, tokens that early-exit at earlier layers contribute the overall reduction rate more than the other tokens, which results in enhanced efficiency. On the other hand, tokens in DeeBERT and PABEE mostly early-exit at the last quarter of the layers owing to one token-determined early-exiting.\par

\section{Ablation Study} \label{ablation study}
\subsection{Training on Different Thresholds}
\label{supsec:Training on Different Thresholds}
The inputs to the Hidden State Mapper (HSM), consisting of intermediate hidden states, are dynamically determined based on the pre-set early-exit threshold. This threshold significantly influences the saturation level of the inputs—lower thresholds result in less saturated inputs, providing more opportunity for the HSM to enhance their quality, and higher thresholds yield more stabilized inputs, requiring less adjustment. Consequently, the training quality of the HSM depends on the specific threshold used during training, making it an essential hyperparameter to consider.

\begin{figure}[ht]
\begin{center}
\centerline{\includegraphics[width=0.65\columnwidth]{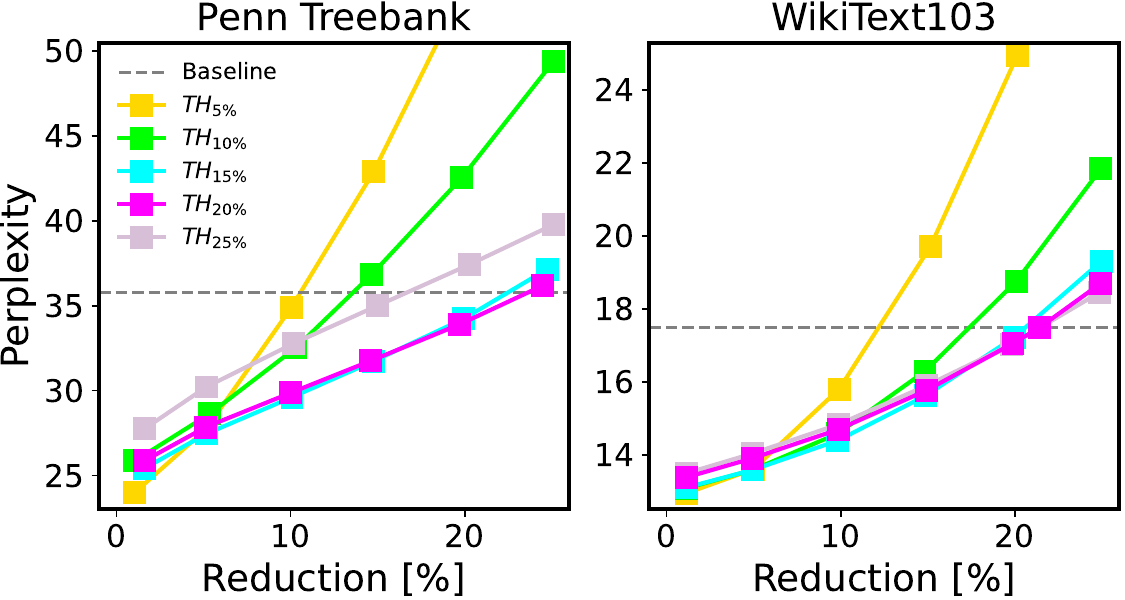}}
\caption{Performance of ADEPT for the GPT2 X-Large model trained by 5 different thresholds for Penn Treebank and WikiText103.}
\label{different-threshold}
\end{center}
\vspace{-5mm}
\end{figure}

To empirically study the effect of threshold on ADEPT's performance, we trained the HSM on GPT2 XL and Llama 7B models with 5 different thresholds for PTB and WikiText103 benchmarks. The thresholds are picked based on the resultant reduction rates (5\%-25\%). The results of GPT2 XL are presented in Figure~\ref{different-threshold}. In both benchmarks, thresholds for the smaller reduction lead to better performance on the smaller reduction, while thresholds for larger reduction enable ADEPT to show the same performance as the base model with the larger reduction. We interpret that a threshold for larger reduction exposes the HSM to a wider range of hidden state refinement during training, which offers more options in terms of reduction rate showing the same or better perplexity than the base model. In the case of a threshold for smaller reduction, the module is more optimized for a narrower range of hidden states in terms of saturation, which enables ADEPT to perform better at a target or smaller reduction rate than cases trained by thresholds for larger reduction. According to the results, we utilized the threshold for 20\% reduction for training in experiments of GPT2 models.\par

\begin{figure*}[!ht]
\begin{center}
\centerline{\includegraphics[scale=0.45]{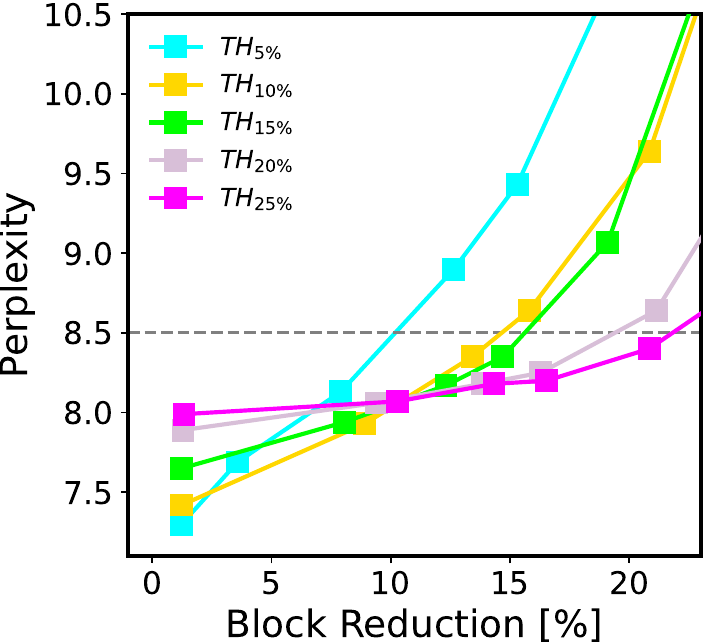}}
\caption{ADEPT for the Llama 7B model trained by different thresholds for WikiText103.}
\label{different-threshold-llama}
\end{center}
\end{figure*}

Figure~\ref{different-threshold-llama} shows the results of Llam 7B model with the 5 different thresholds for WikiText103 benchmark.
Similar to the results from GPT2 XL case, smaller thresholds lead to better performance on the lower reduction, while larger thresholds enable ADEPT to show the same performance as the base model with the larger reduction. According to these results, we utilized the threshold for 25\% reduction for training Llama models.\par

\subsection{All-at-Once Early-Exit}
\label{supsec:All-at-Once Early-Exit}
\begin{figure}[!ht]
\begin{center}
\centerline{\includegraphics[width=0.75\columnwidth]{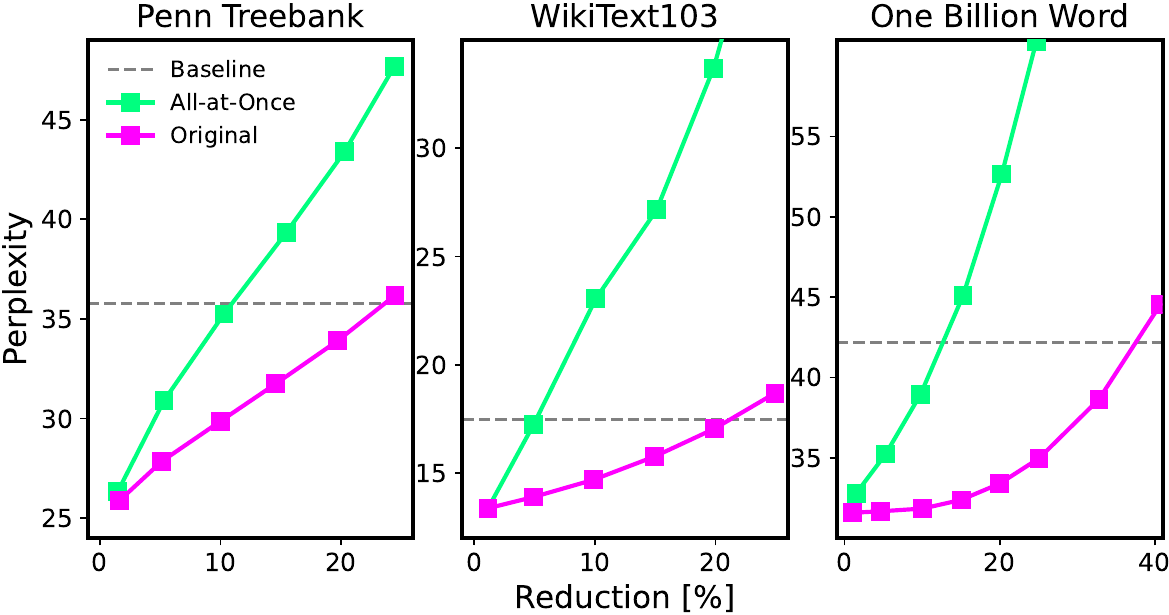}}
\caption{Comparison of original ADEPT and all-at-once ADEPT with GPT2 X-Large model on 3 benchmarks.}
\label{fig:all-at-once}
\end{center}
\vspace{-6mm}
\end{figure}
To evaluate the impact of token-by-token early-exit during the prefill phase on ADEPT's performance, we compare the original ADEPT with an alternative approach, "all-at-once ADEPT." In this variant, early-exit decisions during the prefill phase are based solely on the last token of the input prompt, causing all tokens to exit simultaneously. While the exiting mechanism mirrors previous methods, other components, such as the similarity metric and the add-on module, remain identical to the original ADEPT.\par

The experimental results for the GPT2 XL model are shown in Figure~\ref{fig:all-at-once}. Across all benchmarks, the original ADEPT (labeled as "Original" in the figure) consistently outperforms the all-at-once approach, demonstrating the significant advantage of ADEPT's dynamic, token-level early-exit strategy.

\subsection{Local vs Global Metric}
\label{supsec:Local vs Global Metric}

\begin{figure}[ht]
\begin{center}
\centerline{\includegraphics[width=0.7\columnwidth]{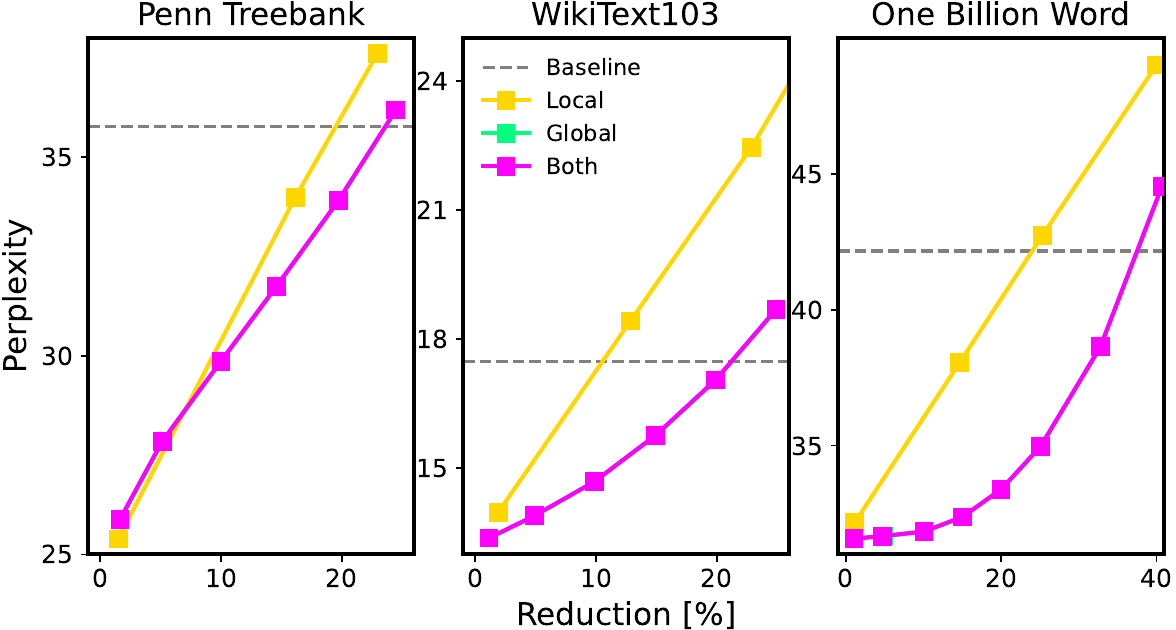}}
\caption{Comparison of utility of local and global metrics of ADEPT with GPT2 X-Large on 3 benchmarks.}
\label{fig:local-vs-global metric}
\end{center}
\vspace{-5mm}
\end{figure}
The global saturation of hidden states is utilized as the early-exit metric in ADEPT, ensuring that the Hidden State Mapper (HSM) receives stabilized intermediate hidden states that are closer to the original final hidden states. If stabilization is measured only locally—by comparing the current hidden state (\(s_{ij}^h\)) with the previous hidden state (\(s_{(i-1)j}^h\))—there is a risk of premature exit caused by temporary stabilization. The importance of global saturation is demonstrated through a comparison of ADEPT's performance using the local metric, global metric, and a combination of both, as shown in Figure~\ref{fig:local-vs-global metric}. Notably, the performance difference between using only the global metric and combining both local and global metrics is negligible, as the "global" case overlaps with the "both" case in the figure.

Across all datasets, the global metric consistently improves the performance of GPT2 XL with ADEPT compared to the local metric.

\section{Details of Token-level Early-Exit Process} \label{sec:details of token-level early-exit}

\begin{figure}[ht]
\begin{center}
\centerline{\includegraphics[scale=0.65]{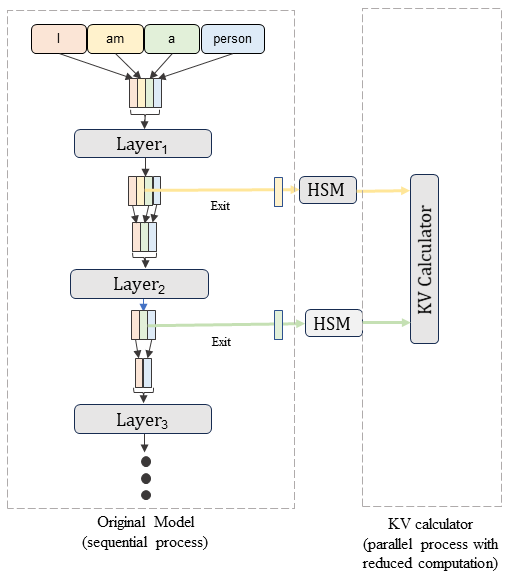}}
\caption{Illustration of token-level early-exit in the prefill phase.}
\label{fig:token-level EE in batch}
\end{center}
\end{figure}

Figure~\ref{fig:token-level EE in batch} illustrates the token-level early-exit process in ADEPT during the prefill phase, applied to an example input prompt, "I am a person." For simplicity, we assume the prompt is tokenized into \textless I\textgreater, \textless am\textgreater, \textless a\textgreater, and \textless person\textgreater, resulting in a token tensor of shape \([4, D_h]\), where \(D_h\) represents the dimensionality of the hidden states. Tokens are processed in parallel across the decoder layers of a large language model (LLM), similar to batch processing. When processing \(N\) prompts in a batch, the tensor shape becomes \([4N, D_h]\). The figure depicts a scenario where tokens \textless am\textgreater (yellow) and \textless a\textgreater (green) early-exit after decoder layers 1 and 2, respectively.

Initially, all tokens are processed through decoder layer 1, as the token \textless am\textgreater only exits after this layer. The token tensor remains \([4, D_h]\). After the early-exit of \textless am\textgreater, the remaining tokens are passed to decoder layer 2, reducing the tensor shape to \([3, D_h]\). After decoder layer 2, the token \textless a\textgreater exits, further shrinking the tensor to \([2, D_h]\).

The tensor size (or batch size) dynamically reduces as tokens exit early, with adjustments performed at each decoder layer. This dynamic shrinkage follows the update:
\[
s^h[\text{Indices}_{\text{active}}] = \textbf{Decoder}(s^h[\text{Indices}_{\text{active}}]),
\]
where \(\text{Indices}_{\text{active}}\) denotes the indices of tokens that have not yet exited. These indices are updated at every layer based on the early-exit policy described in Section~\ref{sec:token-level early-exit}, ensuring that only non-exited tokens are processed by the decoder. This approach directly reduces the computational cost of decoder layers, lowering latency and accelerating the processing of input tokens.

Tokens that early-exit are separately handled to compute key-value (\(s^{kv}\)) representations for skipped layers, as shown in Figure~\ref{fig:token-level EE in batch}. Unlike regular decoder layers—which involve multi-head attention with QKV generation, softmax, linear transformations, and feedforward computations—the early-exiting tokens bypass these computations. Instead, their key-value representations are generated by a streamlined KV Calculator. This reduction in operations for early-exited tokens further decreases latency and improves overall efficiency.

In the summarization phase, as depicted in Figure~\ref{fig:token-level EE in batch}, LLMs operate sequentially, generating one token at a time and feeding it back to the model. For example, tokens \textless with\textgreater and \textless happiness\textgreater are sequentially generated and early-exit after decoder layers 2 and 3, respectively. In this case, decoder layers 1–2 are executed for \textless with\textgreater, and layers 1–3 are executed for \textless happiness\textgreater. While the base model without early-exit processes all tokens through all layers, resulting in a sequence length of 10 (5 tokens × 2 decoder layers each), ADEPT reduces the effective sequence length to 5, significantly accelerating the process.

The early-exit process for the prefill phase, illustrated in Figure~\ref{fig:token-level EE in batch}, follows the same principles depicted for the summarization phase in Figure~\ref{fig:token-level EE in batch}. Notably, key-value generation for skipped layers (Process 2) is computationally lightweight compared to the full decoder operations (Process 1). For instance, in GPT2 XL, key-value computations require approximately 12 times less computation than regular decoder processing, as discussed in Section~\ref{sec:KV GEN}. As a result, latency due to key-value generation is negligible, ensuring that ADEPT's improvements in efficiency remain dominant. 

\subsection{Hidden State Mapper Architecture}
The network architecture of the Hidden State Mapper (HSM), illustrated in Figure~\ref{fig:hsm}, was determined based on cost-effectiveness after evaluating multiple design alternatives. Increasing the MLP size beyond the chosen configuration yielded negligible performance improvements. To analyze the role of the first convolutional block, we tested alternative configurations on the GPT2 X-Large model using the Penn Treebank (PTB) dataset, as shown in Table~\ref{tab:hsm_architecture_options}. 

In the first alternative, the convolutional block was removed, and the input to the HSM was reshaped from \((3, D_h)\) to \((3 \cdot D_h)\), requiring adjustments to the first fully connected (FC) layer. In the second alternative, the convolutional block was replaced with an element-wise addition operation, which reshaped the input from \((3, D_h)\) to \((D_h)\). Both alternatives showed worse performance compared to the original design, with the first alternative also resulting in a larger model size. These findings highlight the effectiveness of the convolutional block in the HSM architecture.

\begin{table}[!ht]
\caption{Performance comparison of different HSM architectures on GPT2 XL model using the Penn Treebank dataset. Metrics include reduction in computation (\%) and perplexity (lower is better).}
\label{tab:hsm_architecture_options}
\begin{center}
\resizebox{0.7\columnwidth}{!}{%
\begin{tabular}{lcc}
\toprule
\textbf{HSM Architecture} & \textbf{Reduction [\%]} $\uparrow$ & \textbf{Perplexity} $\downarrow$ \\
\midrule
Reshaping Inputs \& Changing the 1st FC Layer & 10 & 29.9 \\
                                              & 20 & 34.7 \\
Element-wise Addition                         & 10 & 31.7 \\
                                              & 20 & 37.8 \\
\textbf{Original HSM Structure (Baseline)}    & \textbf{10} & \textbf{29.9} \\
                                              & \textbf{20} & \textbf{33.9} \\
\bottomrule
\end{tabular}%
}
\end{center}
\end{table}

\clearpage

\end{document}